%% file: root.tex
\documentclass[letterpaper, 10 pt, conference]{ieeeconf}  

\IEEEoverridecommandlockouts                              

\input{preamble.tex}
\usepackage[utf8]{inputenc}
\usepackage{commath}
\usepackage{mathtools}
\usepackage{mathabx}
\usepackage{siunitx}
\usepackage[ruled,vlined]{algorithm2e}
\usepackage[lofdepth,lotdepth]{subfig}
\usepackage[font=small]{caption}

\usepackage[belowskip=-10pt,aboveskip=0pt]{caption}
\graphicspath{{media/}}

\newcommand{\ie}{i.e., }

\acrodef{lidar}{Light Detection And Ranging}
\acrodefplural{lidar}{Light Detection And Ranging}
\acrodef{ICP}{Iterative Closest Point}
\acrodef{GNSS}{Global Navigation Satellite System}
\acrodef{IMU}{Inertial Measurement Unit}
\acrodef{DOF}{Degrees Of Freedom}
\acrodef{RTK}{Real Time Kinematic}
\acrodef{RTS}{Robotic Total Station}
\acrodefplural{RTS}{Robotic Total Stations}
\acrodef{SLAM}{Simultaneous Localization and Mapping}
\acrodef{DOF}{Degrees Of Freedom}
\acrodef{MTM}{Modified Transverse Mercator}

\title{\LARGE \bf
Benchmarking ground truth trajectories with robotic total stations
}

\author{Effie Daum$^{{1,2}}$, Maxime Vaidis$^{1}$ and François Pomerleau$^1$
\thanks{$^{1}$Northern Robotics Laboratory, Université Laval, Québec City, Québec, Canada, $\{$maxime.vaidis, francois.pomerleau$\}$@norlab.ulaval.ca }
\thanks{$^{2}$ École Supérieure des Géomètres-Topographes, Le Mans, France, effie.daum.auditeur@lecnam.net}
}

\begin{document}

\maketitle


\begin{abstract}
Benchmarks stand as vital cornerstones in elevating \ac{SLAM} algorithms within mobile robotics.
Consequently, ensuring accurate and reproducible ground truth generation is vital for fair evaluation.
A majority of outdoor ground truths are generated by \ac{GNSS}, which can lead to discrepancies over time, especially in covered areas.
However, research showed that \ac{RTS} setups are more precise and can alternatively be used to generate these ground truths.
In our work, we compare both \ac{RTS} and \ac{GNSS} systems' precision and repeatability through a set of experiments conducted weeks and months apart in the same area.
We demonstrated that \ac{RTS} setups give more reproducible results, with disparities having a median value of \SI{8.6}{\milli\meter} compared to a median value of \SI{10.6}{\centi\meter} coming from a \ac{GNSS} setup.
These results highlight that \ac{RTS} can be considered to benchmark process for \ac{SLAM} algorithms with higher precision.
\end{abstract}


\section{Introduction}

\begin{figure}[thbp]
	\centering
	\includegraphics[width=\linewidth, trim={0 0 0 0}, clip]{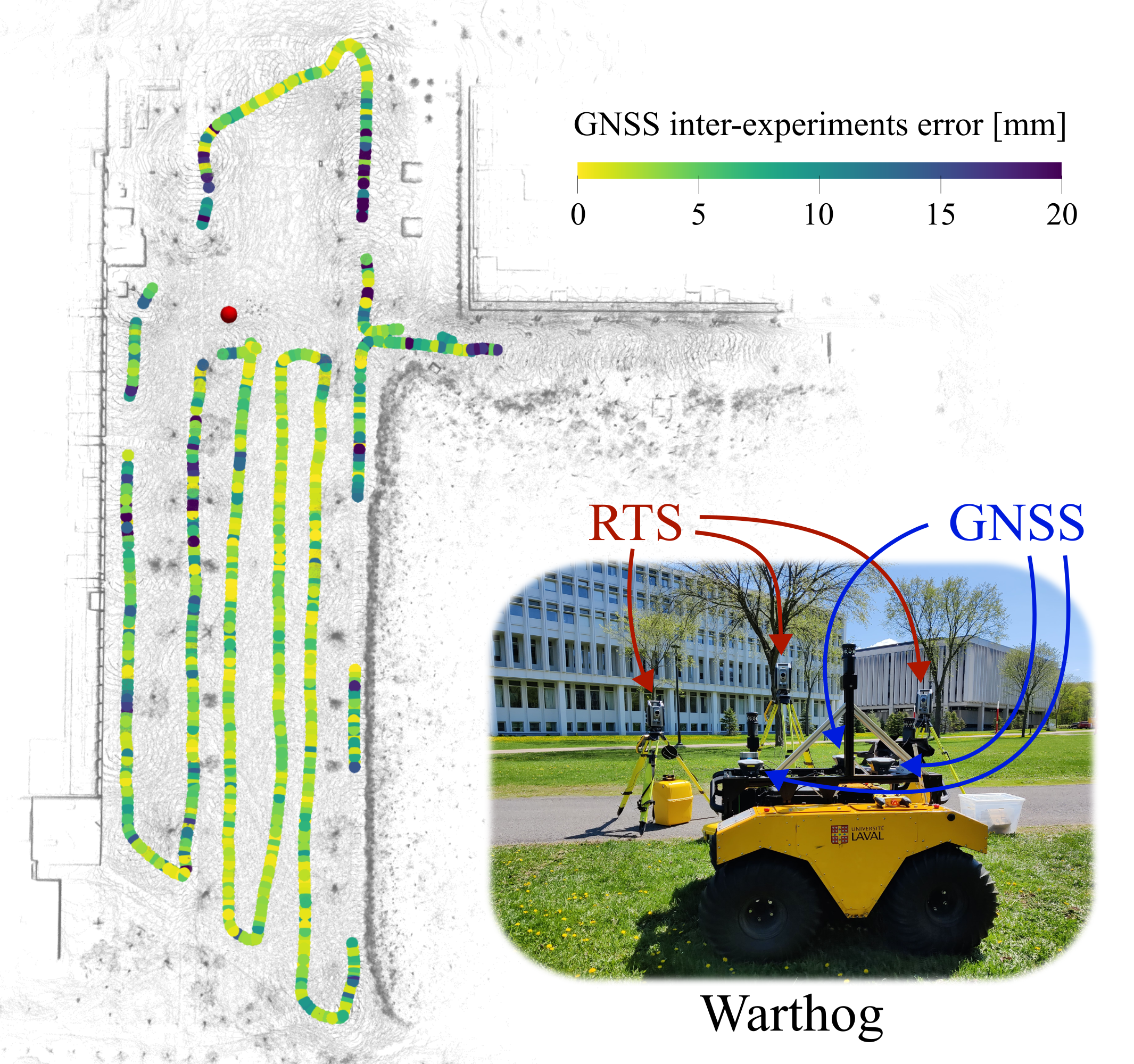}
	\caption{
 A \ac{RTS} setup and \ac{GNSS} antennas were used to record the trajectory of a \textit{Warthog Clearpath} platform on the Université Laval campus.
 The color bar displays the average \acs{GNSS} disparities obtained between two identical trajectories done at different times.
 The red sphere marks the location of our static \ac{RTK} \ac{GNSS} reference antenna.
 }
	\label{fig:intro}
\end{figure}

Benchmarks play a crucial role in enhancing \ac{SLAM} algorithms and real-time location algorithms in mobile robotics~\cite{Euroc2016, helmberger2022hilti}.
It is essential to ensure the accuracy and reproducibility of the ground truth used for fair comparisons between evaluated algorithms~\cite{Maset2022}.
However, outdoor ground truths, primarily generated by \ac{GNSS}, can lead to disparities between experiments conducted at different times in the same environment, as shown in~\autoref{fig:intro}.
These variations in \ac{GNSS} positions result from various sources, such as satellite constellations, ephemerid, and atmospheric conditions. 
They may cause significant biases when evaluating trajectories through benchmarks~\cite{kitti}. 
Recently, our research has demonstrated that \ac{RTS} can generate ground truths in six-\ac{DOF} with millimeter-level accuracy~\cite{Vaidis2021, vaidis2023iros}. 
Building on these findings, we evaluate the feasibility of using \acp{RTS} to generate ground truth trajectories for objective benchmarking of \ac{SLAM} algorithms. 
We compare the precision and repeatability of a \ac{RTS} and \ac{GNSS} system by analyzing their data taken simultaneously during different deployments.
\section{Benchmarking Standardized Experiments}

\subsection{Standardization of \ac{RTS} and \ac{GNSS} protocol}

The experiments were conducted following a standardized protocol to ensure accurate and reproducible results. 
The process began by allowing the \ac{RTS} surveying instruments to acclimate to the outdoor temperature. 
Once ready, the instrument was leveled to ensure proper alignment.
Three prisms were mounted at different heights on the robotic platform to optimize visibility and tracking.
Each prism was associated with its respective \ac{RTS}, also positioned at different heights to avoid obstructing visibility and to facilitate the extrinsic calibration of the sensors.
An essential aspect of the experiment is the extrinsic calibration, where a set of eight to twelve static ground control points is measured around the \ac{RTS} in a circular configuration to express all the data in a common frame~\cite{Vaidis2023icra}.
Finally, after each deployment, we performed a final extrinsic calibration in the laboratory by measuring the positions of prisms and sensors on the robotic platform using a \ac{RTS}.
The same procedure was applied during each experiment to collect consistent and standardized data during all deployments.

The data obtained from each deployment were processed using our pipeline.\footnote{\url{https://github.com/norlab-ulaval/RTS_project}}
The pipeline incorporates various filtering techniques to enhance the precision of the ground truth. 
These filtering methods contribute to minimizing noise and errors, ultimately improving the reliability of the generated trajectories.
As \ac{RTS} positions are not taken synchronously, we used the parameters described by \citet{vaidis2023iros} to perform linear interpolation of the positions.
A point-to-point minimization is used to reconstruct the full pose of the vehicle by measuring three rigid points~\cite{Vaidis2021}.
By leveraging this comprehensive processing pipeline and utilizing the same parameters, the experiment aimed to achieve more precise results, thus facilitating the evaluation and validation of the robotic platform's localization and mapping performance.

Additionally, three \ac{GNSS} antennas are mounted on top of the prisms to achieve precise positioning. 
A fourth static antenna, located nearby with known global geodesic coordinates, provides real-time corrections to the three mobile antennas on the robotic platform for \ac{RTK}-positioning.
The \ac{RTK} method allows obtaining real-time measurements for the trajectory of the moving platform.
Establishing a radio connection between the static antenna and the three mobile antennas, along with setting mask parameters, is crucial for the system's \ac{GNSS} method. 
By using the same point-to-point minimization method as for the \ac{RTS} solution~\cite{Vaidis2021}, the robot ground truth trajectory can be determined in the \ac{GNSS} frame, through the extrinsic calibration of the \ac{GNSS} antennas done in the laboratory.

\subsection{Metrics}
\label{subsec:metrics}
An inter-distance metric is used to evaluate the precision of each system.
This metric is computed with the distance between each synchronous triplet of \ac{RTS} target position (inter-prism distances) or \ac{GNSS} antenna position (inter-\ac{GNSS} distances) obtained during an experiment.
Each of these distance triplets is then compared to their \ac{RTS} calibrated distance, \ie the position of the prisms or \ac{GNSS} antennas rigidly installed on the robot.
Moreover, an inter-experiment metric is used to quantify the difference in precision obtained between two experiments done at different times.
Two positions taken during different experiments are assessed to be in close range by computing their nearest neighbor distance. 
Then, each inter-distance triplet of the \ac{RTS} prisms or \ac{GNSS} antenna positions
that matched spatially are subtracted to compute this metric.
The results represent the disparities in precision in-between the two different trajectories taken at a different time, as shown in~\autoref{fig:intro} for a set of \ac{GNSS} data.
\section{Results}
%
The \ac{RTS} setup is composed of three \textit{Trimble} S7 surveying instruments that track three \textit{Trimble MultiTrack Active Target} MT1000 prisms, operating at a measurement rate of \SI{2.5}{\hertz}. 
The prisms are mounted on a \textit{Clearpath Warthog} unmanned ground vehicle, along with three \textit{Emlid RS+} \ac{GNSS} antennas.
To analyze the disparities of the different setups, eleven experiments were conducted weeks and months apart on the same area of the Université Laval campus, for a total of \SI{16}{\kilo\metre} of \ac{GNSS} and \ac{RTS}-tracked prism trajectories.

\begin{figure}[thbp]
	\centering
 	\includegraphics[width=\linewidth, trim={0 0 0 5}, clip]{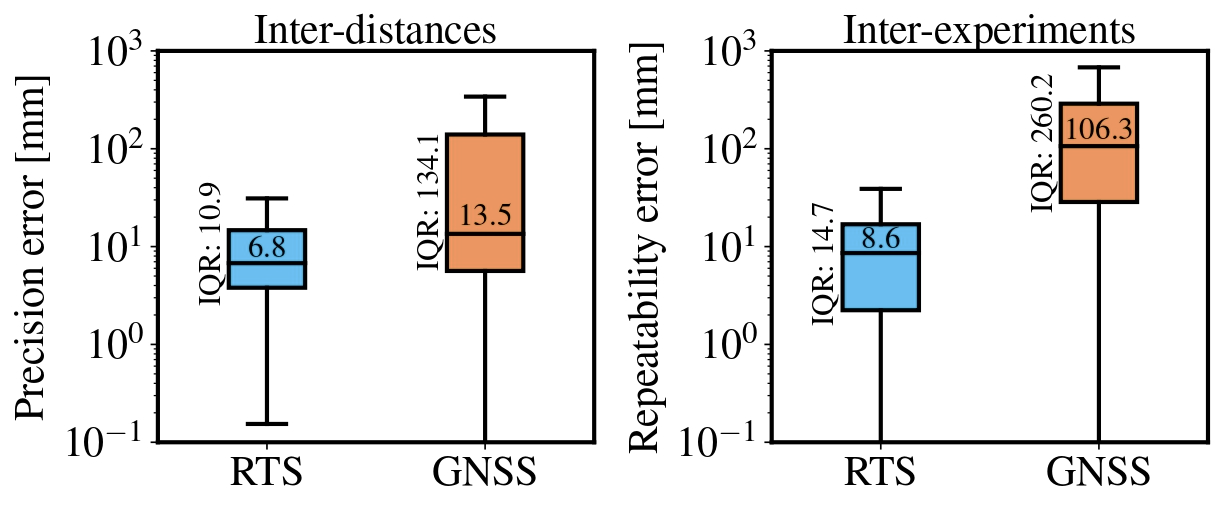}
	\caption{
    Error resulting from \textbf{(a)} inter-prisms and inter-\acp{GNSS} metrics and, \textbf{(b)} inter-experiments metrics presented in~\autoref{subsec:metrics}.
    The results from the \ac{RTS} are depicted in blue, while those from the \ac{GNSS} are represented in orange. 
    The median error is displayed at the center of each box, and the Interquartile Range (IQR) is depicted on the side.}
	\label{fig:results}
\end{figure}

%
%
The \autoref{fig:results} illustrates the inter-prism and inter-\ac{GNSS} metric errors, indicating that the \ac{RTS} acquisition system achieves median sub-centimeter precision at \SI{6.8}{\milli\meter}, while the \ac{GNSS} system provides a median precision around \SI{1.35}{\cm}.
The GNSS precision aligns with results from an \ac{RTK} method, showing within \SI{2}{\centi\meter} accuracy in static scenarios~\cite{morales2007dgps}. 
These outcomes are especially promising considering the dynamic nature of the robotic platform.
It's worth noting that the inter-distances error highlights the higher precision of the \ac{RTS} acquisition system compared to the \ac{GNSS} system. 
This discrepancy can be attributed to the relatively low error of the \ac{RTS} acquisition system versus the absolute error of \ac{GNSS} related to the satellites' constellation, even in open-sky and large-space environments where the experiments were done.

%
%
Reproducibility between the experiments is assessed by computing the nearest neighbor distance. 
Points falling within a \SI{2}{\meter} range are considered reproducible between the different experiments. 
As evident in~\autoref{fig:results}, the reproducibility appears consistent for the \ac{RTS} setup.
This showcases that precision remains consistent across all experiments with a median margin of \SI{8.6}{\milli\meter}.
However, the \ac{GNSS} system has higher disparities at a median level of \SI{10.6}{\centi\meter}.
The ground truth trajectory generated is displayed in~\autoref{fig:intro} with the color gradient showing the inter-experiments error and grey points representing the \ac{SLAM} generated map.
These differences highlight that \ac{GNSS} is less prompt to give reproducible ground truth trajectories with lower uncertainty in the test conditions.


\section{Conclusion}
In this paper, we successfully integrated both \ac{RTS} and \ac{GNSS} ground truth acquisition systems for trajectory reconstruction. 
\acp{RTS} offers a valuable solution for benchmarking due to their higher precision, their median reproducibility around \SI{8.6}{\milli\meter}, and applicability as shown in~\autoref{fig:results}.
Moreover, they can be used in both indoor and outdoor environments compared to \ac{GNSS}.
Results for \ac{GNSS} are as expected, with higher disparities at a median level of \SI{10.6}{\centi\meter}, making it a relevant subsidiary to obtain reproducible trajectories.
However, it's important to note that \ac{RTS} has certain limitations, such as line of sight dependency, higher cost, and post-processing requirements. 
Despite these drawbacks, combining both \ac{RTS} and \ac{GNSS} systems presents a favorable trade-off. 
This approach enables us to generate accurate ground truth trajectories and enhance reproducibility, thereby improving the overall benchmarking process for \ac{SLAM} algorithms.
Future works should consider the complementarity of both systems for six-\ac{DOF} trajectory reconstruction.

%
\renewcommand*{\bibfont}{\small} 
\printbibliography
\end{document}

%% file: preamble.tex
\usepackage[l2tabu,orthodox]{nag}

\usepackage[
backend=biber, 
bibencoding=utf8,
style=ieee, 
sorting=none, 
natbib=true, 
doi=false, 
isbn=false, 
url=false, 
eprint=false, 
maxcitenames=1, 
mincitenames=1,
]{biblatex}

\usepackage[pdftex, hidelinks, colorlinks, hyperfootnotes=false]{hyperref} 

\usepackage[dvipsnames]{xcolor} 

\usepackage[printonlyused]{acronym}

\usepackage{siunitx}
\usepackage[all]{nowidow}

\usepackage[dvipsnames]{xcolor}

\usepackage{lipsum}

\usepackage[pdftex]{graphicx}

\usepackage{epstopdf}

\usepackage{import}

\graphicspath{{./latexGoodPractices/}}

\usepackage{booktabs}

\usepackage{tabu}

\usepackage{amssymb,amsfonts,amsmath,amscd}

\usepackage{bm} 

\newcommand{\bbm}{\begin{bmatrix}}
\newcommand{\ebm}{\end{bmatrix}}